# CNN Is All You Need

**Qiming Chen, Ren Wu**

{qchen,renw}@novumind.com

NovuMind Inc, USA

**Abstract**

The Convolution Neural Network (CNN) has demonstrated the unique advantage in audio, image and text learning; recently it has also challenged Recurrent Neural Networks (RNNs) with long short-term memory cells (LSTM) in sequence-to-sequence learning, since the computations involved in CNN are easily parallelizable whereas those involved in RNN are mostly sequential, leading to a performance bottleneck. However, unlike RNN, the native CNN lacks the history sensitivity required for sequence transformation; therefore enhancing the sequential order awareness, or position-sensitivity, becomes the key to make CNN the general deep learning model. In this work we introduce an extended CNN model with strengthen position-sensitivity, called PoseNet. A notable feature of PoseNet is the asymmetric treatment of position information in the encoder and the decoder. Experiments shows that PoseNet allows us to improve the accuracy of CNN based sequence-to-sequence learning significantly, achieving around 33-36 BLEU scores on the WMT 2014 English-to-German translation task, and around 44-46 BLEU scores on the English-to-French translation task.

## 1   Introduction

CNNs have been successfully used in audio, image and text classification, analysis and generation [12,17,18], whereas the RNNs with LSTM cells [5,6] have been widely adopted for solving sequence transduction problems such as language modeling and machine translation [19,3,5]. The RNN models typically align the element positions of the input and output sequences to steps in computation time for generating the sequenced hidden states, with each depending on the current element and the previous hidden state. Such operations are inherently sequential which precludes parallelization and becomes the performance bottleneck. This situation has motivated researchers to extend the easily parallelizable CNN models for more efficient sequence-to-sequence mapping. Once such efforts can deliver satisfactory quality, the usage of CNN in deep learning would be significantly broadened.

Compared to the history-sensitive recurrent models, in sequence-to-sequence learning, convolution models provide the means for efficient non-local referencing across time steps without fully sequential processing, allowing the computations over the whole sequence to be concurrent rather than one element at a time, and further maximizing GPU's capability for orders of magnitude performance gain.

Convolution is generally conducted by the matrix operations on batches of records. To capture the element-wise sequential context in the record-level processing, several mechanisms have been proposed recently such as position encoding, kernel dilation, attention, etc, in ConvS2S [7], Xception [6], ByteNet [12], WaveNet [17] and SliceNet [11]. To reduce the number of parameters, and hence the computation cost, some additional optimizations such as depth-wise convolution [11] and multi-head attention [19], have been introduced, which is commonly characterized by dividing elements along the channel dimension, parallelizing the sub-processing and then concatenating the partial results.

In this work we focus on further enhancing the position-sensitivity in the CNN based sequence-to-sequence learning framework, we explore how to distinguish various operational phases to apply the right mechanisms in the right contexts, for maximizing the expected benefits and minimizing the unwanted side-effects. Specifically we encapsulate a group of neurons for convolutions in a convolution box whose activity vector represents the instantiation parameters – a similar treatment found in [7,11], as the basic building blocks of our architecture. When using such convolution boxes in the encoder and the decoder, we customize their internal structures depending on where they are used, with different sub-layers for dealing with the sequential position information in the corresponding context. A notable feature of our approach is the asymmetric treatment of position information in the encoder and the decoder. According to the difference of how sequential information is presented and used in the encoding and the decoding processes, we repeat position-encoding (or timing signal) along multiple layers only in the encoder; apply dilation convolution for encoding and regular convolutions for decoding, and organize self-attention, cross-attention, position-wise feed-forward with customized residual links [8], selectively in both the encoder and the decoder.

We implement the above context-sensitive position-sensitive mechanisms in an extended CNN model, PoseNet, for improving the accuracy of convolution based sequence-to-sequence learning. Experiments show that using PoseNet allows us to achieve around 33-36 approximate BLEU scores on the WMT 2014 English-to-German translation task with batch-size 2048 and 250K-500K training steps, as well as to get superior performance in English-to-French translation, achieving 44-46 approximate BLEU scores with batch-size 2048 and 1000K training steps.

In the similar way we also enhanced the accuracy of the "attention-only" approach [19] by an extended model that actually outperforms the PoseNet described here in the same tasks; however, as we believe that in certain areas such as image recognition, the CNN can provide higher generality, in this work we focus on exploring the universal use of the CNN, particularly in sequence-to-sequence learning, and have the other efforts reported separately.

## 2  History-Sensitivity and Position-Sensitivity

Sequence-to-sequence learning is typically based on the encoder-decoder architectures [10,11,13,19,21] where the encoder processes an input sequence $x = (x_1, \ldots, x_n)$ of $n$ elements and returns the internal representations $h = (h_1, \ldots, h_n)$, and the decoder takes $h$ to generate the output sequence $y = (y_1, \ldots, y_m)$ left to right, one element at a time.

In the RNN based sequence-to-sequence learning [5,9], the above $h$ is computed sequentially and kept as the revisable, long or short, history. To generate output $y_{i+1}$, the decoder computes a new hidden state $h_{i+1}$ based on the previous state $h_i$, the previous output $y_i$, as well as a conditional input $c_i$ derived from the encoder output $h$. The models without attention consider only the final encoder state $h_n$, either by ignoring $c_i$ or by setting $c_i = h_n$ for all position $i$. Architectures with attention compute $c_i$ as a weighted sum of $(h_1, \ldots, h_n)$ at each time-step as the corresponding attention scores, focusing on different parts of the input sequence. Attention scores are computed by comparing each encoder state to a combination of the previous decoder state and the last prediction element; the result is normalized to be a distribution over input elements [1].

The CNN based sequence-to-sequence transformation follows this high-level scenario in general, but with the intermediate encoder and decoder representations calculated by convolutions in parallel for all input and output positions. Usually both encoder and decoder networks are stacked with a kind of convolution layer that computes intermediate representations based on a fixed number of input elements. Stacking several layers on top of each other increases the range of input elements represented. Further, convolutions are often followed by non-linearities, allowing the networks to focus on wider input field. Padding is employed in both encoding and decoding to ensure the match of the input length, and to ensure that at a step, no future information is available to the decoder.

In summary, in sequence-to-sequence learning, RNN relies on history sensitive sequential computation, but CNN relies on position-sensitive parallel computation.

## 3 Strengthen Position-Sensitivity in Convolutional Sequence Learning

To optimize CNN for sequence-to-sequence learning, let us first understand how the concept of "sequence" is caught in a CNN framework.

In sequence transduction tasks, the sense of "sequence" is represented by long-range dependencies. One important factor affecting the ability to learn such dependencies is the length of the paths between any combination of positions in the input and output sequences, which the signals have to traverse forward and backward in the network; the shorter these paths the easier to learn long-range dependencies.

From this point of view convolution provides a key advantage for sequence-to-sequence learning, since a multi-layer convolution stack can create multi-level representations over the input sequence where nearby input elements interact at lower levels and distant elements interact at higher levels [7]. This way, the hierarchical structure modeled by CNN provides shorter paths compared to the chain structure modeled by RNN. In the CNN with kernels of width $k$, a feature representation capturing relationships within a window of $n$ elements (such as words) can be accessed by applying only $O(n/k)$ convolution operation, compared to a linear number $O(n)$ in an RNN.

Further, the CNN based sequence mapping can be naturally parallelized since inputs are fed through a constant number of kernels and non-linearities, whereas the number of operations and non-linearities applied in an RNN varies from position to position.

### 3.1 How Position Relationships Captured

Convolution, like other deep learning operations, is essentially a tensor-to-tensor mapping. In sequence transformations, the positional relationships between the elements are handled along with the matrix manipulation of tensors.

**Position Encoding**

In sequence-to-sequence learning, an input or target record consists of a sequence of elements along time-steps. A common way for a deep learning model to make use of the order of the elements is to inject some information about the relative or absolute position of the elements in the sequence [7]. Typically the input elements $x = (x_1, \ldots, x_m)$ are first embedded in distributional space $w = (w_1, \ldots, w_m)$ with depth $d$. To equip the model with a sense of the absolute position of input elements, $p = (p_1, \ldots, p_m)$ is encoded, forming the combined input element representation $e = (w_1+p_1, \ldots, w_m+p_m)$. The target elements are processed similarly in their encoding phase. Note that the positional encodings form a tensor with the same depth $d$ as the input embeddings, so that the two can be summed. Usually positional encodings are added to the input embeddings at the bottoms of the encoder and decoder stacks.

There exist various choices of positional encodings; the following position encoding function $f_{pe}$ uses **sine** and **cosine** functions of different frequencies:

$$f_{pe}(pos, 2i) = sin(pos/10000^{2i/d})$$

$$f_{pe}(pos, 2i+1) = cos(pos/10000^{2i/d})$$

where $pos$ is the position, $i$ the dimension and $d$ the depth.

Since position encoding gives the model a sense of order of the input or target sequence it is currently dealing with, we explored the way to repeat it at the appropriate layers of the model graph to strengthen the position sensitivity at multiple layers, but without introducing unexpected noises.

**Position-wise Feed-Forward Networks (ffn)**

Encoding/decoding in sequence-to-sequence mapping essentially consists in determining the positional correlation between the input/target pairs, this is why the position-wise feed-forward networks (*ffn*) come to the picture. An *ffn* is similar to a convolution with kernel size 1 that is applied to each position separately and identically. An *n*-layer $ffn^n$ performs *n* linear transformations with a ReLU activation in between, which can be intuitively described as below.

$$ffn^1(x) = xW_1 + b_1$$

$$ffn^2(x) = max(0, xW_1 + b_1)W_2 + b_2$$

$$ffn^n(x) = max(0, ffn^{n-1}(x))W_n + b_n$$

Although such linear transformations are the same across different positions, it is position awareness due to that the transformations use different parameters from layer to layer. Therefore adding *ffn* to the appropriate points in the model graph provides a way to enhance position sensitivity.

**Filter Dilation**

Filter dilation is a mechanism for correlating distant elements in convolutional sequence-to-sequence autoregressive approach. Essentially, filter dilation increases the receptive field of the convolution operation by enlarging the spatial extent from which feature information can be gathered. In the other words, the dilated convolution operators can use the same filter at different ranges using different dilation factors.

Compared to using larger convolution windows, using filter dilation has the pros of lower computation cost, and the cons of unequal convolutional coverage of the input space. Our observation indicates that for boosting position awareness, the dilation mechanism has stronger effect in encoder than in decoder.

**Cross-Attention and Self-Attention**

The simple inner-product attention correlates two tensors position-wise. Given two tensors $S[n, d]$ and $T[m, d]$, where *d* stands for depth, according to [11,19], the attention mechanism computes the feature vector similarities at each position and re-scales according to the depth:

$$attention(S, T) = 1/\sqrt{d} \cdot Softmax(T \cdot S^T) \cdot S$$

where *S* and *T* can be two different tensors or the same tensor; we refer to the attention in the former case cross-attention, and in the latter case self-attention.

Cross-attention is often used in the "encoder-decoder attention" layers, where *T* comes from the previous decoder layer, and *S* from the encoded input, i.e. output of the encoder. This allows every position in the decoder to attend over all positions in the input sequence. Self-attention is used in input encoding and target encoding where *S* and *T* come from the same place – the output of the previous layer in the encoder or in the decoder. A self-attention layer in the encoder allows each position to attend to all positions in the previous layer of the encoder; similarly, a self attention layer in the decoder allow each position to attend to all positions in the decoder up to and including that position. To prevent leftward information flow in the decoder to preserve the auto-regressive property [11,12,17], masking out the values in the input of the softmax which correspond to illegal positions, is necessary.

One way to strengthen the positional relationship in convolution based sequence-to-sequence learning is to apply the cross-attention and self-attention multiple times. We have experienced this with certain accuracy gain.

### 3.2 Design Consideration

In order to capture richer sequential position information and position-wise relationships between inputs and targets for improved accuracy of CNN based sequence-to-sequence learning, we follow these design considerations: the impact of position-sensitivity on the accuracy of sequence-to-sequence learning varies in the encoding process and decoding process; as a result, context specific convolution boxes are needed for applying the above position-sensitive mechanisms.

Intuitively, when inputs are encoded, the sequences dealt with are to be completely populated thus sensitive to enhanced (e.g. repeated) position encoding. However, during decoding, in the partially generated targets with pads as space-fillers, some of the above mechanisms, such as repeated position encoding, would be ineffective or even noisy. The context sensitivity of other mechanisms for capturing position information can be explained similarly.

## 4  PoseNet Architecture

Our PoseNet architecture is built using the *tensor2tensor* library [23] and extending the Slicenet [11] model. We follow the conceptual encoder-decoder structure [10,11,13,19,21], where the encoder maps an input sequence representations $(x_1, ..., x_n)$ to a sequence of continuous hidden representations $h = (h_1, ..., h_n)$; from there, the decoder then generates an output sequence $(y_1, ..., y_m)$ one element at a time. At each step the model is auto-regressive, consuming the previously generated elements as additional input when generating the next. Our architecture realizes this overall encoder-decoder structure using stacked convolution boxes and point-wise, fully connected layers, shown in the left and right halves of **Figure 1**, respectively. However, these convolution boxes are customized differently for the encoder and the decoder.

### Encoder

The encoder is composed of a stack of L = 6 layers. Each layer has two convolution boxes with residual links and a simple, position-wise fully connected feed-forward net. For each sub-layer with function *f*, we employ a residual connection followed by layer normalization [1], i.e. produce *norm(x + f(x))*. In addition, each layer begins with a position encoding, and ends with a position-wise feed-forward net, both of these allow us to strengthen the position-sensitivity of our model as they are repeated in all the L = 6 layers. We also choose to invoke the dilation convolutions provided in the *tensor2tensor* library for encoding, to position-wise correlate distant elements. It is worth noting that we repeat position-encoding and use dilation convolution only for encoding, but not for decoding, for the reasons explained.

### Decoder

Our model follows the convolutional autoregressive structure introduced in Slicenet[11], ByteNet [12], WaveNet [17] and PixelCNN [18]. Inputs are embedded and encoded before being fed into a decoder that auto-regressively generates each element of the output. At every step, the autoregressive decoder produces a new output prediction given the encoded inputs and the encoding of the existing predicted outputs. The basic modules are the convolution boxes stacked, and the attention modules for the decoder to get information from the encoder. As we reuse the corresponding functions from the *tensorflow* library and the *tensor2tensor* library, we skip their details here.

We use a stack of L = 5 decoder layers. The cross-attention, that allows each decoding step to attend the encoded-inputs, is repeatedly applied in each layer. In the same way as the encoder, each decoder layer ends with a position-wise feed-forward net. Each decoder layer also has two convolution boxes with residual links and a simple, position-wise fully connected feed-forward network. Masking is used to prevent positions from attending to subsequent positions. The mask and the offset of output embeddings ensure that the prediction for each position to depend only on the known outputs at positions less than that position.

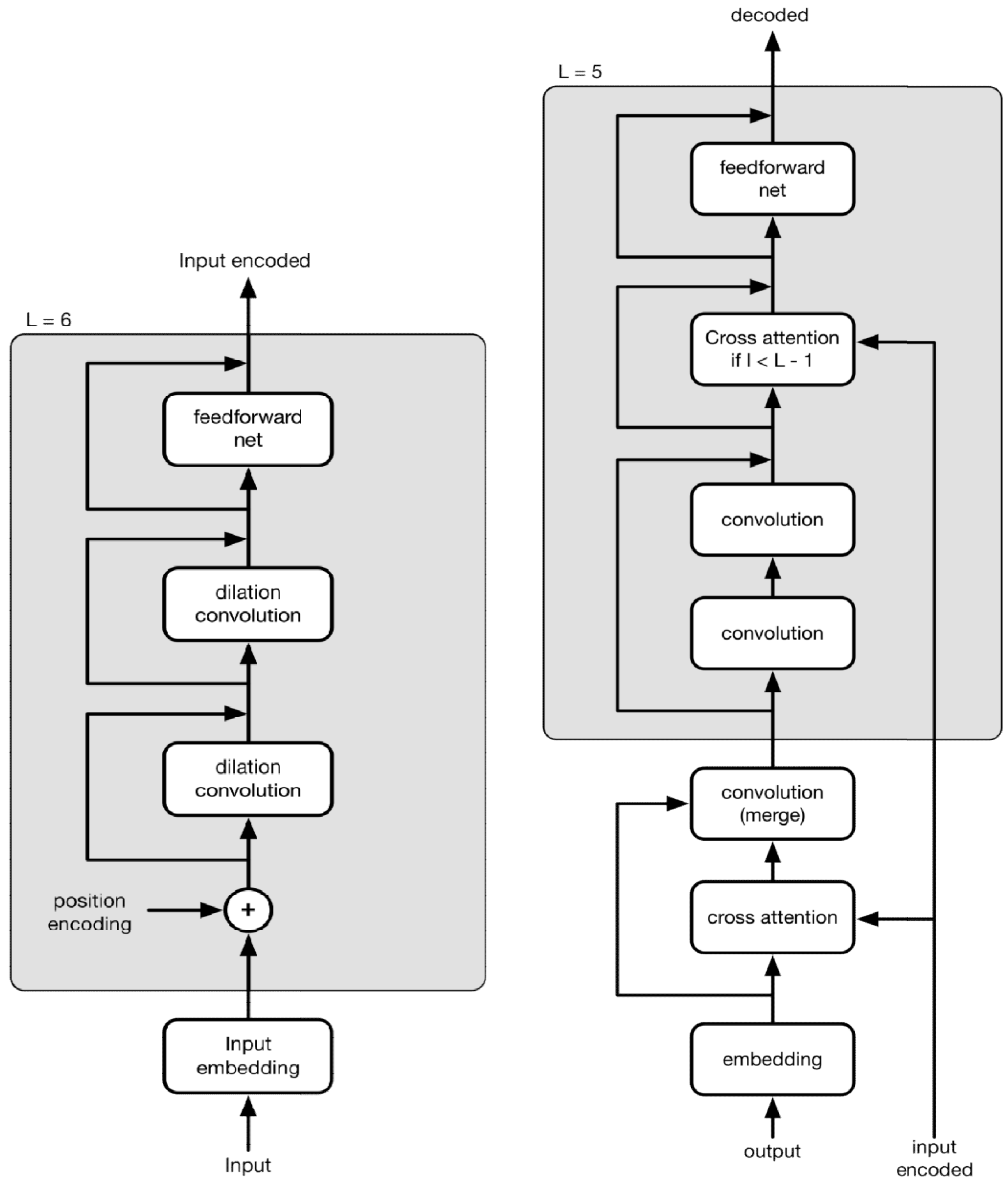

Figure 1: The encoder (left) and decoder (right) stacks:

In summary, in capturing the sequential position information and the position-wise relationships between inputs and targets, the PoseNet treats the convolution boxes for the encoder and those for the decoder differently; in the other words, the encoder and the decoder employ context specific convolution boxes which involve the position-sensitive mechanisms in different ways. In the encoder stack, the position encoding and the position-wise feed-forward net are applied to the beginning and the ending of each layer, repeatedly, and the convolutions are dilated to be more sensitivity to the distant positions. In the decoder stacks, the cross-attention is used at each layer repeatedly, but the position encoding is optionally applied only once.

We chose not to use dilated convolution for decoder as in the step-by-step decoding process, outputs and pads shift thus less sensitive to dilation. For the same reason, we do not apply, or optionally apply only once without repeating in every layer, the position encoding, because in this case it has less effect or even adds noise. Please also note that in each layer the cross-attention is place before the convolution boxes.

For enhanced parallelism and/or reduced parameters, the mechanisms of multi-head attention [19] and depth-wise convolution [11] are introduced, which are in common characterized by splitting the channels of an input into several non-overlapping segments, performing a regular attention or spatial convolution over each segment independently and then concatenating the resulting feature maps along the channel axis. The details of these mechanisms can be found in [19], [11] and the *tensorflow* API specifications. In this work we take advantage of them.

## 5   Experiments

We carry out our experiments on x86_64 GNU/Linux with 4.4.0-96-generic #119-Ubuntu with 8G memory, using one NVIDIA GeForce GTX 1070 with CUDA V8.0.61, python3.5.2, tensorflow-1.3.0, tensor2tensor-1.2.2. In this study we focus on the follows issues:

- How to improve the position sensitivity in the CNN-based, non-recurrent sequence-to-sequence learning for enhanced accuracy, and
- How to apply the position-sensitivity strengthen mechanisms differently in the encoder and in the decoder.

Our model is trained and evaluated for the WMT English-to-German (EN-DE) and English-to-French (EN-FR) translation tasks using the benchmark data *translate_ende_wmt32k* and *translate_enfr_wmt32k*. The sizes of the sample data, measured on disk, are about 689M for EN-DE and 11G for EN-FR. All of our experiments are implemented using the *tensorflow* framework and the *tensor2tensor* library [23]. We also leveraged the convolution and attention mechanisms described in the related work [4, 11, 12,17, 19].

Our experiments reveal that
- the convolution based sequence-to-sequence learning can be benefited by the combination of several position sensitivity strengthen mechanisms such as repeatedly imposing timing signal, selectively apply dilation, appropriate mix parameterized and convolution-based self-attention, etc;
- certain position sensitivity strengthen mechanisms are more effective for accuracy enhancement if applied to encoding, including input encoding and target encoding, rather than to decoding.

The intermediate training states are checkpointed by the *tensorflow* framework; along with each checkpoint (e.g. 2000 steps) the corresponding evaluation results are logged. We attached the partial evaluation results at the last 10 checkpoints on training the EN-DE translation model in Appendix 8.1; and on training the EN-FR translation model in Appendix 8.2. The average of these BLEU scores are approximately consistent with those measured at the 10 additional checkpoints after the given train steps. We did not average the BLEU scores measured at the previous checkpoints where the model is still under-trained.

Using our model, with batch-size 2048 and using 1 GPU, for the EN-DE translation task we achieved 33-36 approximate BLEU scores in 250K-500K training steps; for the EN-FR translation task we achieved 44-46 approximate BLEU scores in 1000K training steps (on a multiple GPU machine we see slightly higher accuracy but lower BLEU scores).

To justify the significance of our approach, we have the BLEU scores on the EN-DE and the EN-FR translations compared with the related work (given in [11,19]), as listed in **table 1** below.

**WMT 2014 En-De Translation Task**

| Model | EN-DE BLEU | EN-FR BLEU |
|---|---|---|
| Bytenet [12] | 23.75 | |
| GNMT + RL [20] | 24.6 | 39.92 |
| ConvS2S [7] | 25.16 | 40.46 |
| MoE [14] | 26.03 | 40.56 |
| GNMT + RL Ensemble [22] | 26.30 | 41.16 |
| ConvS2S Ensemble [7] | 26.36 | 41.29 |
| Transformer (base) [19] | 27.3 | 38.1 |
| Transformer (big) [19] | 28.4 | 41.0 |
| SliceNet (Full, 2048) [11] | 25.5 | |
| SliceNet (Super 2/3, 3072) [11] | 26.1 | |
| **PoseNet (2048)** | **33-36 (avg 34.92)** (250K-500K steps) | **44-46 (avg 45.54)** (1000K steps) |

**Table 1**: Performance of our models in EN-DE and EN-FR translation tasks with benchmark data *translate_ende_wmt32k* and *translate_enfr_wmt32k* compared to the latest published results; the "avg BLEU" averages the scores measured at the last 10 checkpoints

# 6 Conclusions

The goal of providing a generalized and efficient deep-learning framework has motivated us to explore the possibility of using CNN as the universal building blocks. As CNN already has the superior track-record in audio, image and text learning, we particularly focus on its potential in sequence-to-sequence learning. Compared to the computations involved in RNN, the computations involved in CNN are easily parallelizable, but less history sensitive. Therefore enhancing the sequential order awareness, or position-sensitivity, is the key for CNN to support sequence transformation. In this work we introduce a CNN architecture, PoseNet, which is characterized by applying the position-sensitive mechanisms - position encoding (or timing signal), self-attention and cross-attention, dilation convolution, position-wise feed-forward net, residual net, etc, selectively in the encoder and the decoder for maximizing their effects. A notable feature of PoseNet is the asymmetric treatment of position information in the encoder and the decoder. For this we turn the common convolution boxes into specific ones depending on where they are used, with different sub-layers for capturing the context specific sequence oriented information. Experiments show that with strengthen position-sensitivity, PoseNet is capable of improving the accuracy of convolutional sequence-to-sequence learning - achieving around 33-36 approximate BLEU scores, and 44-46 approximate BLEU scores, on the WMT 2014 English-to-German and English-to-French translation tasks respectively.

# 8 Appendices

## 8.1 PoseNet on translate_ende_wmt32k: Evaluation at last 10 checkpoints of 500K steps

```
INFO:tensorflow:Finished evaluation at 2017-10-21-11:57:14
INFO:tensorflow:Validation (step 482000):
metrics-translate_ende_wmt32k/neg_log_perplexity = -2.01664,
loss = 1.82994, metrics-translate_ende_wmt32k/rouge_2_fscore = 0.41238,
metrics-translate_ende_wmt32k/accuracy = 0.631383,
metrics-translate_ende_wmt32k/rouge_L_fscore = 0.606513,
metrics-translate_ende_wmt32k/approx_bleu_score = 0.336328,
metrics-translate_ende_wmt32k/accuracy_top5 = 0.811289

INFO:tensorflow:Finished evaluation at 2017-10-21-12:25:11
INFO:tensorflow:Validation (step 484000):
metrics-translate_ende_wmt32k/neg_log_perplexity = -1.95574,
loss = 1.73249, metrics-translate_ende_wmt32k/rouge_2_fscore = 0.421828,
metrics-translate_ende_wmt32k/accuracy = 0.639055,
metrics-translate_ende_wmt32k/rouge_L_fscore = 0.617276,
metrics-translate_ende_wmt32k/approx_bleu_score = 0.342415,
metrics-translate_ende_wmt32k/accuracy_top5 = 0.820492

INFO:tensorflow:Finished evaluation at 2017-10-21-12:53:18
INFO:tensorflow:Validation (step 486000):
metrics-translate_ende_wmt32k/neg_log_perplexity = -1.96284,
loss = 1.75151, metrics-translate_ende_wmt32k/rouge_2_fscore = 0.422178,
metrics-translate_ende_wmt32k/accuracy = 0.635866,
metrics-translate_ende_wmt32k/rouge_L_fscore = 0.618148,
metrics-translate_ende_wmt32k/approx_bleu_score = 0.350611,
metrics-translate_ende_wmt32k/accuracy_top5 = 0.821233

INFO:tensorflow:Finished evaluation at 2017-10-21-13:24:14
INFO:tensorflow:Validation (step 488000):
metrics-translate_ende_wmt32k/neg_log_perplexity = -1.93029,
loss = 1.75333, metrics-translate_ende_wmt32k/rouge_2_fscore = 0.417444,
metrics-translate_ende_wmt32k/accuracy = 0.643301,
metrics-translate_ende_wmt32k/rouge_L_fscore = 0.611268,
metrics-translate_ende_wmt32k/approx_bleu_score = 0.34653,
metrics-translate_ende_wmt32k/accuracy_top5 = 0.825383

INFO:tensorflow:Finished evaluation at 2017-10-21-13:52:15
INFO:tensorflow:Validation (step 490000):
metrics-translate_ende_wmt32k/neg_log_perplexity = -1.93131,
loss = 1.73386, metrics-translate_ende_wmt32k/rouge_2_fscore = 0.426887,
metrics-translate_ende_wmt32k/accuracy = 0.643821,
metrics-translate_ende_wmt32k/rouge_L_fscore = 0.61464,
metrics-translate_ende_wmt32k/approx_bleu_score = 0.353239,
metrics-translate_ende_wmt32k/accuracy_top5 = 0.825245

INFO:tensorflow:Finished evaluation at 2017-10-21-14:20:07
INFO:tensorflow:Validation (step 492000):
metrics-translate_ende_wmt32k/neg_log_perplexity = -1.96154,
loss = 1.83409, metrics-translate_ende_wmt32k/rouge_2_fscore = 0.4132,
metrics-translate_ende_wmt32k/accuracy = 0.641521,
metrics-translate_ende_wmt32k/rouge_L_fscore = 0.608894,
metrics-translate_ende_wmt32k/approx_bleu_score = 0.338278,
metrics-translate_ende_wmt32k/accuracy_top5 = 0.821384

INFO:tensorflow:Finished evaluation at 2017-10-21-14:48:09
INFO:tensorflow:Validation (step 494000):
metrics-translate_ende_wmt32k/neg_log_perplexity = -1.97756,
loss = 1.82451, metrics-translate_ende_wmt32k/rouge_2_fscore = 0.413808,
metrics-translate_ende_wmt32k/accuracy = 0.635957,
metrics-translate_ende_wmt32k/rouge_L_fscore = 0.606894,
metrics-translate_ende_wmt32k/approx_bleu_score = 0.344889,
metrics-translate_ende_wmt32k/accuracy_top5 = 0.816165
```

```
INFO:tensorflow:Finished evaluation at 2017-10-21-15:15:54
INFO:tensorflow:Validation (step 496000):
metrics-translate_ende_wmt32k/neg_log_perplexity = -1.90634,
loss = 1.68212, metrics-translate_ende_wmt32k/rouge_2_fscore = 0.44237,
metrics-translate_ende_wmt32k/accuracy = 0.652367,
metrics-translate_ende_wmt32k/rouge_L_fscore = 0.631862,
metrics-translate_ende_wmt32k/approx_bleu_score = 0.367319,
metrics-translate_ende_wmt32k/accuracy_top5 = 0.827552

INFO:tensorflow:Finished evaluation at 2017-10-21-15:44:03
INFO:tensorflow:Validation (step 498000):
metrics-translate_ende_wmt32k/neg_log_perplexity = -1.94244,
loss = 1.73411, metrics-translate_ende_wmt32k/rouge_2_fscore = 0.423232,
metrics-translate_ende_wmt32k/accuracy = 0.639047,
metrics-translate_ende_wmt32k/rouge_L_fscore = 0.61918,
metrics-translate_ende_wmt32k/approx_bleu_score = 0.350548,
metrics-translate_ende_wmt32k/accuracy_top5 = 0.822101

INFO:tensorflow:Finished evaluation at 2017-10-21-16:12:16
INFO:tensorflow:Validation (step 500000):
metrics-translate_ende_wmt32k/neg_log_perplexity = -1.89206,
loss = 1.72012, metrics-translate_ende_wmt32k/rouge_2_fscore = 0.431239,
metrics-translate_ende_wmt32k/accuracy = 0.647935,
metrics-translate_ende_wmt32k/rouge_L_fscore = 0.625391,
metrics-translate_ende_wmt32k/approx_bleu_score = 0.361499,
metrics-translate_ende_wmt32k/accuracy_top5 = 0.828574
```

### 8.2 PoseNet on translate_enfr_wmt32k: Evaluation at last 10 checkpoints of 1000K steps

```
INFO:tensorflow:Finished evaluation at 2017-12-10-15:04:30
INFO:tensorflow:Validation (step 982000):
metrics-translate_enfr_wmt32k/accuracy = 0.675681,
metrics-translate_enfr_wmt32k/approx_bleu_score = 0.450538,
metrics-translate_enfr_wmt32k/accuracy_top5 = 0.850825,
metrics-translate_enfr_wmt32k/rouge_L_fscore = 0.66145,
metrics-translate_enfr_wmt32k/neg_log_perplexity = -1.66226,
metrics-translate_enfr_wmt32k/rouge_2_fscore = 0.50523, loss = 1.37471

INFO:tensorflow:Finished evaluation at 2017-12-10-15:31:52
INFO:tensorflow:Validation (step 984000):
metrics-translate_enfr_wmt32k/accuracy = 0.705628,
metrics-translate_enfr_wmt32k/approx_bleu_score = 0.46891,
metrics-translate_enfr_wmt32k/accuracy_top5 = 0.871277,
metrics-translate_enfr_wmt32k/rouge_L_fscore = 0.678612,
metrics-translate_enfr_wmt32k/neg_log_perplexity = -1.50277,
metrics-translate_enfr_wmt32k/rouge_2_fscore = 0.523873, loss = 1.28076

INFO:tensorflow:Finished evaluation at 2017-12-10-15:59:32
INFO:tensorflow:Validation (step 986000):
metrics-translate_enfr_wmt32k/accuracy = 0.678946,
metrics-translate_enfr_wmt32k/approx_bleu_score = 0.44124,
metrics-translate_enfr_wmt32k/accuracy_top5 = 0.845314,
metrics-translate_enfr_wmt32k/rouge_L_fscore = 0.650633,
metrics-translate_enfr_wmt32k/neg_log_perplexity = -1.66341,
metrics-translate_enfr_wmt32k/rouge_2_fscore = 0.495654, loss = 1.44104

INFO:tensorflow:Finished evaluation at 2017-12-10-16:27:01
INFO:tensorflow:Validation (step 988000):
metrics-translate_enfr_wmt32k/accuracy = 0.690276,
metrics-translate_enfr_wmt32k/approx_bleu_score = 0.458958,
metrics-translate_enfr_wmt32k/accuracy_top5 = 0.860904,
metrics-translate_enfr_wmt32k/rouge_L_fscore = 0.668237,
metrics-translate_enfr_wmt32k/neg_log_perplexity = -1.58758
metrics-translate_enfr_wmt32k/rouge_2_fscore = 0.514411, loss = 1.32547

INFO:tensorflow:Finished evaluation at 2017-12-10-16:54:23
INFO:tensorflow:Validation (step 990000):
```

```
metrics-translate_enfr_wmt32k/accuracy = 0.679814,
metrics-translate_enfr_wmt32k/approx_bleu_score = 0.444471,
metrics-translate_enfr_wmt32k/accuracy_top5 = 0.847369,
metrics-translate_enfr_wmt32k/rouge_L_fscore = 0.646489,
metrics-translate_enfr_wmt32k/neg_log_perplexity = -1.66079,
metrics-translate_enfr_wmt32k/rouge_2_fscore = 0.4958, loss = 1.44457

INFO:tensorflow:Finished evaluation at 2017-12-10-17:21:41
INFO:tensorflow:Validation (step 992000):
metrics-translate_enfr_wmt32k/accuracy = 0.686088,
metrics-translate_enfr_wmt32k/approx_bleu_score = 0.44928,
metrics-translate_enfr_wmt32k/accuracy_top5 = 0.856638,
metrics-translate_enfr_wmt32k/rouge_L_fscore = 0.661313,
metrics-translate_enfr_wmt32k/neg_log_perplexity = -1.60854,
metrics-translate_enfr_wmt32k/rouge_2_fscore = 0.507824, loss = 1.37166

INFO:tensorflow:Finished evaluation at 2017-12-10-17:49:07
INFO:tensorflow:Validation (step 994000):
metrics-translate_enfr_wmt32k/accuracy = 0.677961,
metrics-translate_enfr_wmt32k/approx_bleu_score = 0.450756,
metrics-translate_enfr_wmt32k/accuracy_top5 = 0.853979,
metrics-translate_enfr_wmt32k/rouge_L_fscore = 0.664942,
metrics-translate_enfr_wmt32k/neg_log_perplexity = -1.63439,
metrics-translate_enfr_wmt32k/rouge_2_fscore = 0.501011, loss = 1.35754

INFO:tensorflow:Finished evaluation at 2017-12-10-18:16:43
INFO:tensorflow:Validation (step 996000):
metrics-translate_enfr_wmt32k/accuracy = 0.698592,
metrics-translate_enfr_wmt32k/approx_bleu_score = 0.465756,
metrics-translate_enfr_wmt32k/accuracy_top5 = 0.863186,
metrics-translate_enfr_wmt32k/rouge_L_fscore = 0.675051,
metrics-translate_enfr_wmt32k/neg_log_perplexity = -1.54489,
metrics-translate_enfr_wmt32k/rouge_2_fscore = 0.521121, loss = 1.31394

INFO:tensorflow:Finished evaluation at 2017-12-10-18:44:19
INFO:tensorflow:Validation (step 998000):
metrics-translate_enfr_wmt32k/accuracy = 0.692221,
metrics-translate_enfr_wmt32k/approx_bleu_score = 0.463122,
metrics-translate_enfr_wmt32k/accuracy_top5 = 0.858793,
metrics-translate_enfr_wmt32k/rouge_L_fscore = 0.668282,
metrics-translate_enfr_wmt32k/neg_log_perplexity = -1.59228,
metrics-translate_enfr_wmt32k/rouge_2_fscore = 0.519492, loss = 1.30747

INFO:tensorflow:Finished evaluation at 2017-12-10-19:11:39
INFO:tensorflow:Validation (step 1000000):
metrics-translate_enfr_wmt32k/accuracy = 0.698182,
metrics-translate_enfr_wmt32k/approx_bleu_score = 0.46112,
metrics-translate_enfr_wmt32k/accuracy_top5 = 0.865655,
metrics-translate_enfr_wmt32k/rouge_L_fscore = 0.664928,
metrics-translate_enfr_wmt32k/neg_log_perplexity = -1.51432,
metrics-translate_enfr_wmt32k/rouge_2_fscore = 0.521016, loss = 1.31765
```